%% file: main.tex
\documentclass[runningheads]{llncs}
\usepackage{graphicx}
\usepackage{comment}
\usepackage{amsmath,amssymb} %
\usepackage{color}

\usepackage{booktabs}

\usepackage{microtype}

\def\etal{\emph{et al.} }
\def\ie{\emph{i.e.} }

\usepackage{tabularx}

\newcolumntype{Y}{>{\centering\arraybackslash}X}

\begin{document}
\pagestyle{headings}
\mainmatter
\def\ECCVSubNumber{1263}  %

\title{Uncertainty-Aware Weakly Supervised Action Detection from Untrimmed Videos} %

\titlerunning{Uncertainty-Aware Weakly Supervised Action Detection}
\author{Anurag Arnab 
	\and
	Chen Sun %
	\and
	Arsha Nagrani %
	\and
	Cordelia Schmid %
}
\authorrunning{A. Arnab et al.}
\institute{Google Research \\
\email{\{aarnab, chensun, anagrani, cordelias\}@google.com}\
}
\maketitle

\input{text/abstract}
\input{text/introduction}

\input{text/related_work}

\input{text/method}

\input{text/experiments}

\input{text/conclusion}

\bibliographystyle{splncs04}
\bibliography{bibliography}
\end{document}

%% file: text/abstract.tex
\begin{abstract}
Despite the recent advances in video classification, progress in spatio-temporal action recognition has lagged behind.
A major contributing factor has been the prohibitive cost of annotating videos frame-by-frame.
In this paper, we present a spatio-temporal action recognition model that is trained with only video-level labels, which are significantly easier to annotate. %
Our method leverages per-frame person detectors which have been trained on large image datasets within a Multiple Instance Learning framework.
We show how we can apply our method in cases where the standard Multiple Instance Learning assumption, that each bag contains at least one instance with the specified label, is invalid using a novel probabilistic variant of MIL where we estimate the uncertainty of each prediction.
Furthermore, we report the first weakly-supervised results on the AVA dataset and state-of-the-art results among weakly-supervised methods on UCF101-24.
\keywords{spatio-temporal action recognition, weak supervision, video understanding, mulitple instance learning}
\end{abstract}

%% file: text/introduction.tex
\section{Introduction}

\input{figures/teaser}

Video classification has witnessed great advances recently due to large datasets such as Kinetics~\cite{kay2017kinetics} and Moments in Time~\cite{monfort2018moments} which have enabled training of specialised neural network architectures for video \cite{carreira_cvpr_2017,feichtenhofer_iccv_2019}.
However, progress in other video understanding tasks, such as spatio-temporal action detection, has lagged 
behind in comparison.
There are fewer datasets for action recognition, which are also significantly smaller than their video-classification counterparts.
A reason for this is the exorbitant cost of annotating videos with spatio-temporal labels -- each frame of an action has to be manually labelled with a bounding box.
Moreover, annotating temporal boundaries of actions is not only arduous, but often ambiguous with annotators failing to reach consensus about the start and end times of an action~\cite{cheron_neurips_2018,sigurdsson_cvpr_2017}.

In this paper, we propose a method to train spatio-temporal action detectors using only weak, video-level annotations as shown in Fig.~\ref{fig:teaser}.
To achieve this, we leverage image-based person detectors which have been trained on large image datasets such as Microsoft COCO~\cite{lin_eccv_2014} and are accurate across large variations in appearance, scene and pose.
We adopt a Multiple Instance Learning (MIL) framework, where a person tubelet is an instance, and all person tubelets in the video form a bag.
An important consideration in our approach is the presence of label noise: this is introduced from using off-the-shelf person detectors which have not been trained on the video-dataset of interest, and also the fact that we have to sample tubelets from large bags in long videos due to memory constraints.
In both of these scenarios, the standard Multiple Instance Learning assumption~\cite{diettrich_1997}, that each bag contains at least one instance with the bag-level label, may be violated.
We are not aware of previous work that has explicitly addressed this problem, and we do so with a probabilistic variant of MIL where we estimate the uncertainty of an instance-level prediction. 

Using our approach, we obtain state-of-the-art results among weakly-supervised methods on the UCF101-24 dataset.
Furthermore, we report, to our knowledge, the first weakly-supervised results on the AVA dataset (the only large-scale dataset for spatio-temporal action recognition), where we also show the accuracy trade-offs when annotating video-clips for time intervals of varying durations.

%% file: figures/teaser.tex
\begin{figure}[t]
	\centering
	\includegraphics[trim=0 80 0 20, width=0.98\linewidth]{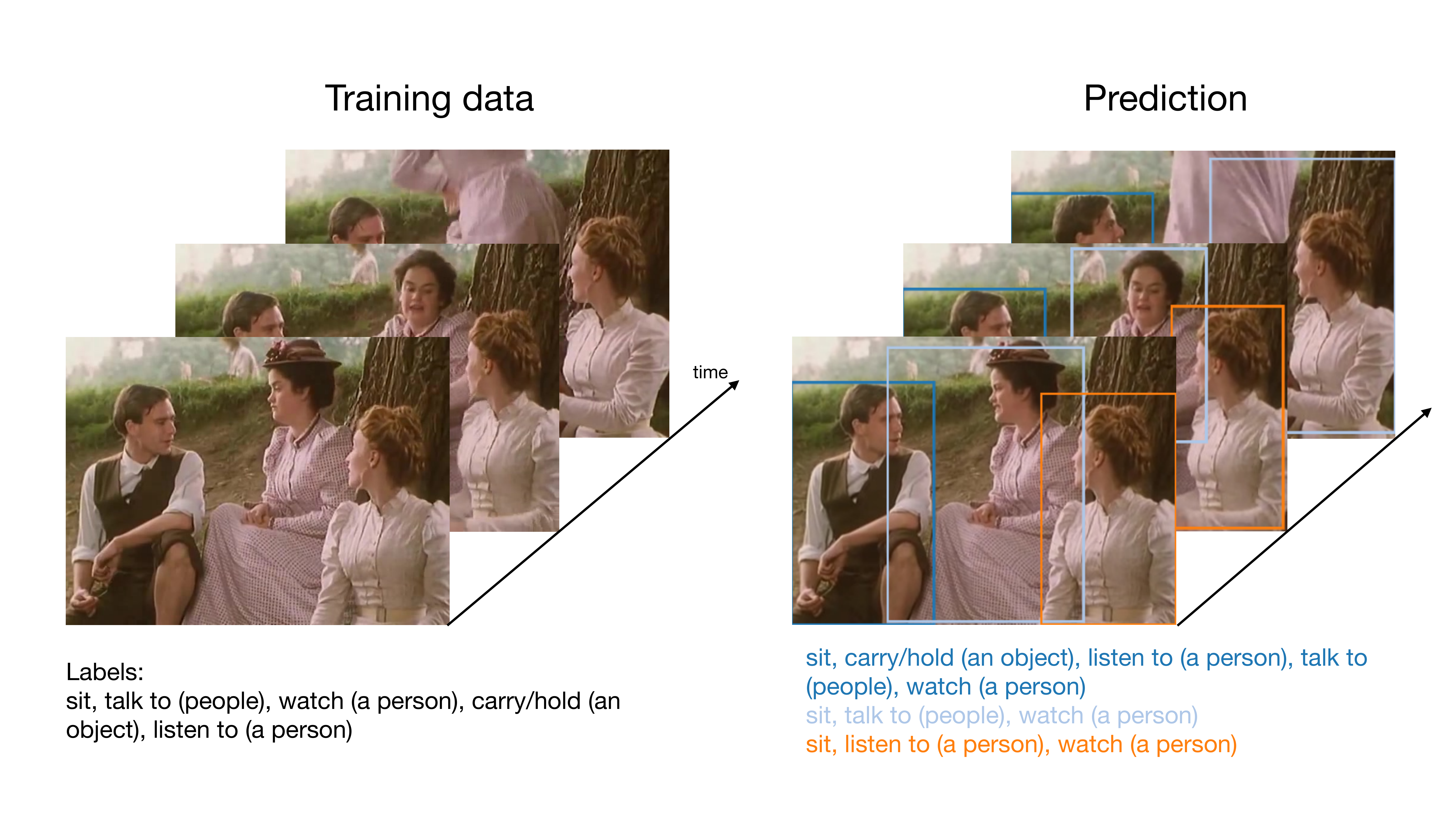}
	\caption{We propose a method to train a spatio-temporal action detector using only weak, video-level labels on challenging, real-world datasets.
	Note that the video-level labels that we have may apply to multiple people in the video, and that these labels may only be active for an unannotated time interval of the input clip.
	}
	\label{fig:teaser}
\end{figure}

%% file: text/related_work.tex
\section{Related Work}

Most prior work on spatio-temporal action recognition has been fully-supervised.
Initial approaches in the area used 3D sliding window detectors in conjunction with handcrafted, volumetric features~\cite{ke_iccv_2005,laptev_iccv_2007}.
Current state-of-the-art approaches are temporal extensions of object detection architectures~\cite{kalogeiton_iccv_2017,singh_iccv_2017,zhao_cvpr_2019,peng_eccv_2016} such as Faster-RCNN~\cite{ren_neurips_2015} and SSD~\cite{liu_eccv_2016}.
These approaches predict bounding boxes around the action in a frame, using as input either a single frame along with optical flow to capture temporal information~\cite{singh_iccv_2017,saha_bmvc_2016} or multiple frames at the input to provide temporal context~\cite{kalogeiton_iccv_2017}.
The predicted bounding boxes are then linked over time using an online, greedy algorithm or dynamic programming to create spatio-temporal tracks.
Our work builds on these methods by also utilising a detection architecture and spatio-temporal linking.
However, these approaches all require bounding box annotations at each frame in the video whilst we only use video-level labels which are significantly cheaper to acquire.

Weakly supervised approaches to spatio-temporal action recognition have also been explored before as they enable a significant reduction in annotation time and cost.
Relevant to our approach is the work of \cite{cheron_neurips_2018}.
Che\'ron \etal \cite{cheron_neurips_2018} also use person detections, and infer their action labels using a formulation based on discriminative clustering~\cite{bach_neurips_2008}.
Although their approach allows them to incorporate different types of supervision, it effectively learns a linear classifier on top of pretrained, deep features.
Our method in contrast is learned fully end-to-end.
Mettes~\etal\cite{mettes2016spot} also employed Multiple Instance Learning (MIL), but used action proposals~\cite{van_gemert_bmvc_2015} instead of the human detections used by our work and \cite{cheron_neurips_2018}.
However, \cite{mettes2016spot}, rely on additional cheap ``point'' annotations (a single spatial point annotated for a subset of the frames which constitute the action) which also ensures that the standard MIL assumption is not violated.
In follow-up work \cite{mettes_bmvc_2017}, the authors removed the need for ``point'' annotations by incorporating biases (\ie the presence of objects in the video, a bias that actions typically occur in the centre of a frame) instead.
Finally, Weinzaepfel \etal\cite{weinzaepfel_arxiv_2016} also used a Multiple Instance Learning framework in conjunction with human detections.
The authors, however, assumed that sparse spatial supervision was present (\ie bounding boxes for a small subset of frames in the action tube), unlike our method which requires video-level labels alone.

We also note that many approaches have addressed temporal action detection (localising actions in time but not space) with only video-level tags as supervision~\cite{wang2017untrimmednets,singh_hide_seek_iccv_2017,nguyen2018weakly,paul2018w}.
UntrimmedNets~\cite{wang2017untrimmednets} uses a network with two branches, a classification module to perform action classification and a selection module to select relevant frames.%
Hide-and-Seek~\cite{singh_hide_seek_iccv_2017} obtains more precise temporal boundaries by forcing the network to attend to more discriminative frames by randomly hiding parts of videos.
However, these methods are trained and evaluated on datasets such as ActivityNet~\cite{caba2015activitynet} and THUMOS14~\cite{jiang2014thumos}, which contain mostly one action per video, and are thus significantly less challenging than datasets such as AVA~\cite{gu2018ava} which we evaluate on.

Finally, we note that another approach to combat the effort of dataset annotation has been various forms of self-supervised learning, where discriminative feature representations can be learned with unlabelled data.
Examples in video include cross-modal self-supervised learning by learning correspondences between the audio and image streams readily available in videos~\cite{arandjelovic2017look,owens2016ambient,zhao2018sound}, transcribed speech~\cite{sun_iccv_2019} or using meta-data such as hashtags~\cite{ghadiyaram2019large} as a form of weak labelling.
Self-supervised approaches, however, are complementary to our approach,  as they still require a limited amount of fully-labelled data for the final task of interest.
In our weakly-supervised action detection scenario, we never have access to full, spatio-temporal ground-truth annotations for a single training example.

%% file: text/method.tex
\section{Proposed Approach}

As shown in Fig.~\ref{fig:teaser}, given a set of video clips, with only clip-level annotations of the actions taking place, our goal is to learn a model to recognise and localise these actions in space and time.
Our method is based on Multiple Instance Learning (MIL) which we briefly review in Sec.~\ref{sec:mil}. Thereafter, we show how we use it for weakly-supervised spatio-temporal action recognition in Sec.~\ref{sec:method_action_recognition_as_mil}.
We then describe how the standard MIL assumption, %
is often violated in our scenario and describe a method to mitigate this by leveraging uncertainty estimates by our network in Sec.~\ref{sec:mil_noise}.
Finally, we discuss implementation details of our network in Sec.~\ref{sec:implementation_details}.

\subsection{Multiple Instance Learning}
\label{sec:mil}

In the standard Multiple Instance Learning (MIL)~\cite{diettrich_1997} formulation, one is given a bag of $N$ instances, denoted as $x = \{x_1, x_2, \ldots, x_N\}$.
The class labels for each of the instances is unknown, but the label for the entire bag, $x$, is known.
The standard MIL assumption is that a bag is assigned a class label if at least one instance in the bag is associated with this label.
More formally, we consider the multi-label classification case, where the label vector for the bag is $y \in \mathbb{R}^{C}$, and $y_l = 1$ if there is at least one instance with the $l^{th}$ label is present in the bag, and $y_l = 0$ otherwise.
Note that each bag can be labelled with multiple of the $C$ class labels.

Our goal is to train an instance-level classifier (parameterised as a neural network), that predicts $p(y_{l} = 1 | x_j)$, or the label probabilities for the $j^{th}$ instance.
However, as we only have the labels for the entire bag, and not each instance, MIL methods aggregate the set of instance-level probabilities, $\{p_{ij}\}$ for a bag $i$, to bag-level probabilities, $p_i$, using an aggregation function, $g(\cdot)$, where the probabilities are obtained from a suitable activation function (sigmoid or softmax) on the logits output by the neural network:
\begin{equation}
p(y_{il} = 1 | x_1, x_2, \ldots, x_N) = g(p_{i1}, p_{i2}, \ldots, p_{iN}).
\label{eq:mil_pooling}
\end{equation}

Once we have bag-level predictions, we can apply a standard classification loss between the bag-level probabilities and bag-level ground truth, and train a neural network with stochastic gradient descent.
Since we consider the multi-label classification case, we use the binary cross-entropy:
\begin{equation}
\mathcal{L}_{ce}(x, y) = -\sum_{i}^{N_b}\sum_{l}^{C} y_{il} \log{p_{il}} + (1 - y_{il})\log(1 - p_{il})
\label{eq:cross_entropy_mil}
\end{equation}
Note that we defined $p_{il}$ as the bag-level probability of the $i^{th}$ bag taking the $l^{th}$ label, which is obtained using Eq.~\ref{eq:mil_pooling}, and $N_b$ is the number of bags in the mini-batch.

\paragraph{Aggregation}
The aggregation function, $g(\cdot)$, can naturally be implemented in neural networks as a global pooling function over all outputs of the network.
Common, permutation-invariant pooling functions include, max-pooling, generalised mean-pooling and log-sum-exponential (LSE) pooling~\cite{boyd_2004} (a smooth and convex approximation of the maximum function) respectively:
\begin{align}
g(\{p_j\}) &= \max_j p_j \\
g(\{p_j\}) &= \left(\frac{1}{\left|j\right|} \sum_j p_j^r\right) ^{\frac{1}{r}} \\
g(\{p_j\}) &= \frac{1}{r} \log \left(\frac{1}{\left|j\right|} \sum_j e^{r \cdot p_j} \right)
\end{align}
Max-pooling only considers the top-scoring instance in the bag, and thus naturally captures the MIL assumption that at least one instance in the bag has the specified, bag-level label.
Moreover, it can also be more robust to instances in the bag that do not have the bag-level label.
However, mean and LSE pooling have been employed in applications such as weakly-supervised 
segmentation \cite{pinheiro_cvpr_2015}, object recognition~\cite{Sun_2016_CVPR} and medical imaging~\cite{kraus_arxiv_2015} where multiple instances in the bag do typically have the bag-level label. 
Note that higher values of the $r$ hyperparameter for both these functions increases their ``peakiness'' and approximates the maximum value.
For our scenario, detailed in the next section, we found max-pooling to be the most appropriate.

\subsection{Weakly-supervised spatio-temporal action recognition as multiple instance learning}
\label{sec:method_action_recognition_as_mil}
\input{figures/method_diagram}

Our goal is to learn a model to recognise and localise actions in space and time given only video-level annotations.
To facilitate this, we leverage a person detector that has been trained on a large image dataset, \ie Microsoft COCO~\cite{lin_eccv_2014}.
Concretely, we run a person detector on our training videos, and create person tubelets which are person detections over $K$ consecutive frames in the video. %
Our bag for multiple instance learning thus consists of all the tubelets within a video, and is annotated with the video-level labels that we have as supervision, as illustrated in Fig.~\ref{fig:method_diagram}.
Note that the size of the bag varies for every video clip, as the bag size is determined by the length of the video and the number of detected people.

As shown in Fig.~\ref{fig:method_diagram}, our network architecture for this task is a Fast-RCNN~\cite{girshick_iccv_2015} style detector that has been extended temporally.
Given a video clip of $K$ frames, and proposals which in our case are person detections, the network classifies the action(s) taking place at the centre frame of each proposal, given the temporal context of the $K-1$ frames around it.

Note that the spatio-temporal localisation task is effectively factorised: the spatial localisation capability of the model depends on the quality of the person detections.
Temporal localisation, on the other hand, is performed by linking person tubelets through the video as commonly done in the literature~\cite{kalogeiton_iccv_2017,singh_iccv_2017,zhao_cvpr_2019,cheron_neurips_2018}, since this method can scale to arbitrarily long videos.
We use the same algorithm as Kalogeiton~\etal\cite{kalogeiton_iccv_2017} which links together detections within a small temporal window greedily based on the spatial intersection over union (IoU) between bounding boxes on consecutive frames.

Finally, note that for a video consisting of $T$ frames, the bag could consist of $T - K + 1$ person tubelets if a person is detected on each frame of the video, and a tubelet is started from each frame.
Due to memory limitations, it is infeasible to fit an entire bag onto a GPU for training.
As a result, we uniformly sample instances from each bag during training, whilst still retaining the original bag-level label.
This introduces additional noise into the problem, as detailed next. %

\subsection{Label noise and violation of the standard MIL assumption}
\label{sec:mil_noise}
\input{figures/loss_surface}

The standard MIL assumption, that at least one instance in the bag is assigned the bag-level label is often violated in our scenario.
There are two primary factors for this:
Firstly, due to computational constraints, we cannot process a whole bag at a time, but must instead sample instances from a bag.
It is therefore possible to sample a bag that does not contain any tubelets with the labelled action.
The likelihood of this occurring is inversely proportional to the ratio of the duration of the labelled action to the total video length.
Secondly, in a weakly-supervised scenario, we use person detectors that are not trained on the video dataset of interest.
Consequently, there can be failures in the detector, especially when there is a large domain gap between the detector's training distribution and the video dataset.
False negatives (missing detections for people in the scene) are a particular issue because it is possible that we do not have a single person tubelet in the bag that corresponds to the labelled action.

Therefore, there are cases when there is no tubelet which actually has the bag-level label.
To handle these cases, inspired by \cite{kendall_neurips_2017,novotny_cvpr_2018}, we modify the network to additionally predict the uncertainty $\sigma \in \ \mathbb{R}^{C}$ for each binary label for all tubelets in the bag.
Intuitively, to minimise the training error, the network can predict the bag-level label with low uncertainty or it can predict a high uncertainty to avoid being penalised heavily for noisy bags where the bag-level label is not present in any of the tubelets.
The final loss, in conjunction with the original cross entropy, is defined as:
\begin{equation}
\mathcal{L}(x, y, \sigma) = \frac{1}{\sigma^2}\mathcal{L}_{ce}(x, y) + \log \sigma^2
\label{eq:uncertainty_loss}
\end{equation}
As shown by \cite{kendall_cvpr_2018}, this corresponds to assuming a Boltzmann distribution on the output of the network with a temperature of $\sigma^2$, and approximately minimising its log-likelihood.

The loss surface of this probabilistic loss is visualised in Fig.~\ref{fig:loss_surface}. %
Note how the loss is the lowest when the predicted label is correct and there is low uncertainty.
However, the loss is not excessive if the incorrect label is predicted with a high uncertainty.
This is in contrast with the standard cross-entropy loss which penalises incorrect predictions heavily.

\subsection{Network architecture and implementation}
\label{sec:implementation_details}

Our action detector is similar to Fast-RCNN~\cite{girshick_iccv_2015} using the SlowFast~\cite{feichtenhofer_iccv_2019} video network architecture based on the ResNet-50 backbone~\cite{he_cvpr_2016} pretrained on Kinetics~\cite{kay2017kinetics}.
As described in Sec.~\ref{sec:method_action_recognition_as_mil}, we use region proposals obtained from a Faster-RCNN detection model trained with Detectron~\cite{Detectron2018}. %
Region-of-interest features~\cite{girshick_iccv_2015} are extracted from the last feature map of ``res5'' using RoIAlign~\cite{he_iccv_2017}.
Our choice for this architecture is motivated by the fact that it is simple and has achieved state-of-the-art results on the AVA dataset~\cite{gu2018ava} in a fully-supervised setting~\cite{feichtenhofer_iccv_2019}.
Note that our network does not use additional optical flow inputs (which can be considered as an additional source of supervision) as common in other video architectures~\cite{carreira_cvpr_2017,kalogeiton_iccv_2017,singh_iccv_2017,cheron_neurips_2018}. 

We predict the uncertainty, $\sigma \in \mathbb{R}^C$ for each of the $C$ binary labels defined by the dataset for each tubelet.
As we use max-pooling to aggregate the tubelet predictions, we select the uncertainty prediction corresponding to the selected tubelet for computing the loss.
For numerical stability, we predict $v := \log \sigma^2$ with our network, using the ``softplus'', $f(x) = \log(1 + \exp(-x))$, activation function to ensure positivity.
We then compute $\frac{1}{\sigma^2} = \exp(-v)$, and avoid the possibility of dividing by 0 which could be the case if we predicted $\sigma^2$ directly with the network.

We train our network with synchronous stochastic gradient descent (SGD), using 8 GPUs and a batch size of 4 on each GPU.
In our case, each element of a batch is of a bag from Multiple Instance Learning.
Each bag samples a maximum of 4 tubelets.
Each tubelet itself consists of 16 frames.

%% file: figures/method_diagram.tex
\begin{figure}[t]
	\centering
	\includegraphics[width=0.98\linewidth]{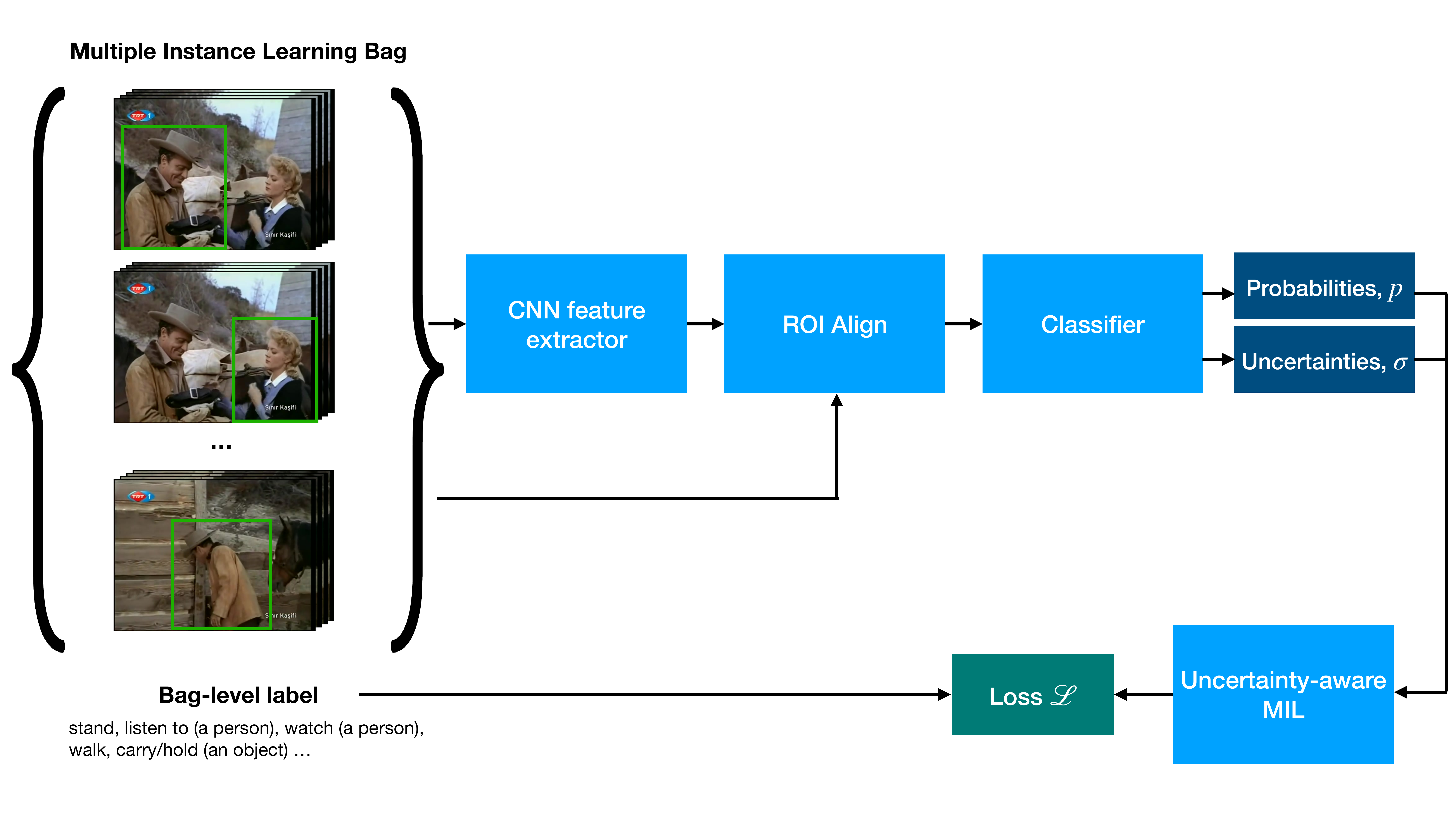}
	\caption{Overview of our approach for training an action detector in a weakly-supervised manner using multiple instance learning: Each bag consists of all the tubelets that have been extracted from the video-clip. These tubelets are obtained using an off-the-shelf person detector which has not been trained on the dataset of interest.
	These tubelets act as proposals for a Fast-RCNN style detector operating on a sequence of rgb images.
	The predictions for each of the tubelets in the bag are then aggregated together, and compared to the bag-level label.
	Uncertainty estimates produced by the network are used to compensate for noise in the bag-level labels during training.
}
	\label{fig:method_diagram}
\end{figure}

%% file: figures/loss_surface.tex
\begin{figure}[tb]
	\vspace{-1\baselineskip}
	\centering
	\includegraphics[trim=0 70 0 30, width=0.90\linewidth]{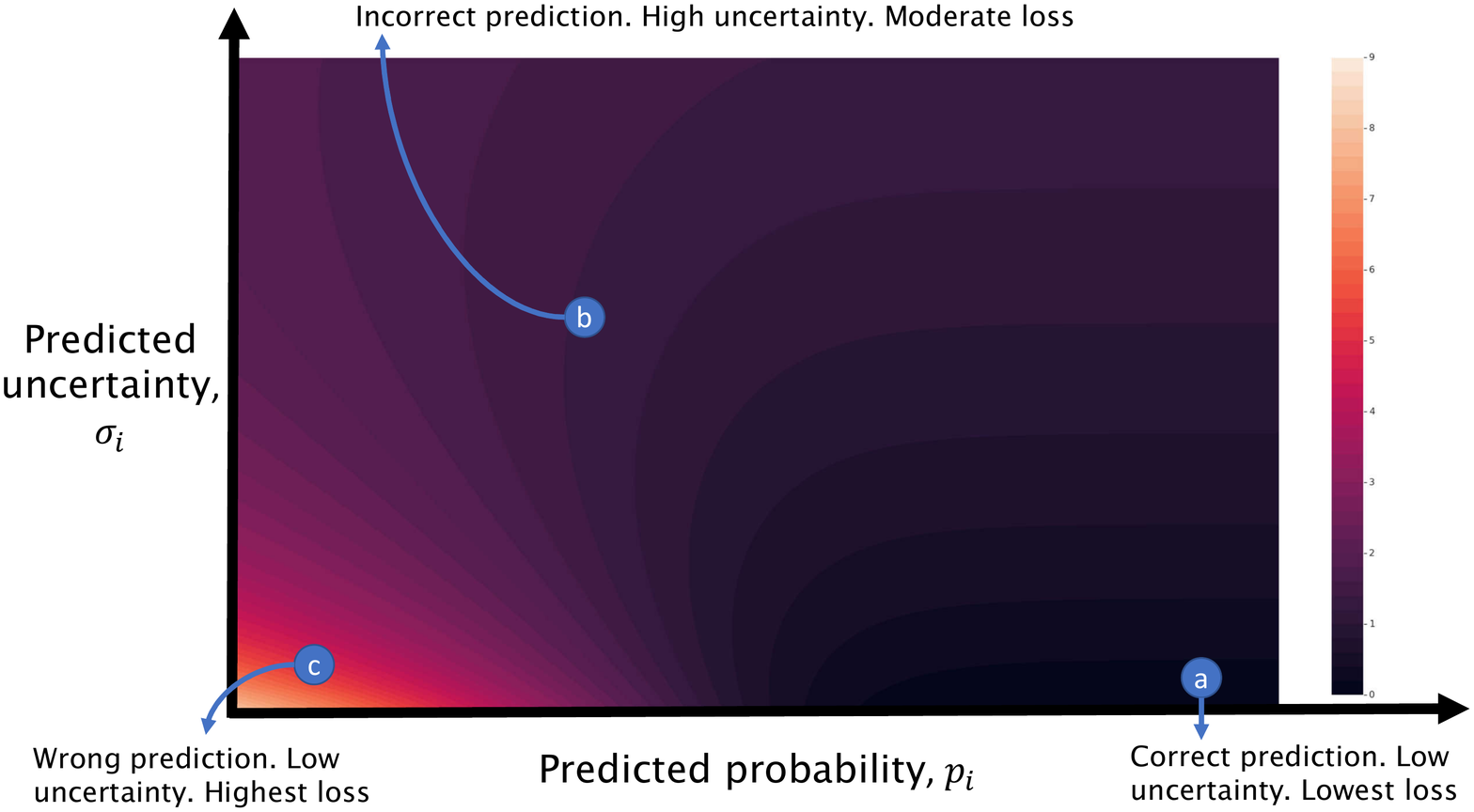}
	\caption{The loss surface of our uncertainty-based loss (Eq.~\ref{eq:uncertainty_loss}).
	The ground truth binary label in this example is 1.
	Hence, the loss is minimised when the network predicts a high probability and a low uncertainty (point ``a'').
	However, making an incorrect prediction with a high uncertainty is not penalised as much (point ``b''), and is suitable for cases when the input bags are noisy and the bag-level label is not present in any of the tubelets.
	Finally, predicting the incorrect label with a low uncertainty is penalised the most (point ``c'').
	Best viewed in colour. 
	}
	\label{fig:loss_surface}
\end{figure}

%% file: text/experiments.tex
\section{Experiments}

\subsection{Experimental set-up}
We evaluate our method on UCF101-24 and AVA, described in more detail below.
Note that other video datasets such as THUMOS~\cite{jiang2014thumos} and ActivityNet~\cite{caba2015activitynet}  are not suitable for spatiotemporal localisation, as they lack bounding box annotations.

\paragraph{UCF101-24:}
UCF101-24 is a subset of the UCF101~\cite{soomro2012ucf101} dataset, consisting of 24 action classes with spatio-temporal localisation annotation, released as bounding box annotations of humans.%
Although each video contains only a single action class, it may contain multiple individuals performing the action with different spatial and temporal boundaries.
Moreover, there may also be people present in the video who are not performing any labelled action.
Following standard practice, we use the corrected annotations of \cite{singh_iccv_2017} and report the mean average precision at a video level (Video AP) for the first split of the dataset.
For evaluating the Video AP, we link tubelets together using the algorithm of \cite{kalogeiton_iccv_2017}.

\paragraph{AVA~\cite{gu2018ava}:}
This dataset consists of 430, 15-minute video clips obtained from movies.
80 atomic visual actions are annotated exhaustively for all people in the video, where one person is often simultaneously performing multiple actions.
The dataset annotates keyframes at every second in the video.
Following standard practice, we report the Frame AP at an IoU threshold of 0.5 using v2.2 annotations.  %

\subsection{Experiments on UCF101-24}

We first conduct ablation studies of our model on the UCF101-24 dataset.
We discard the spatio-temporal annotations for the whole untrimmed video, %
and so our bag in multiple instance learning contains tubelets from the whole video.

\input{tables/ucf_ablation}

\paragraph{Ablation study}
Table~\ref{tab:ucf_ablation} ablates different variants of our method:
The most na\"ive baseline is to not perform any multiple instance learning, and to simply train in a fully-supervised fashion assuming that the label of a tubelet is the video-level label.
As shown in the first row of Tab.~\ref{tab:ucf_ablation}, this method performs the worst as the assumed tubelet-level labels are often incorrect.
The use of multiple instance learning improves results, with the various aggregation functions performing similarly.
Max-pooling, however, performs the best, and we believe this is because the max operation is the most suitable for dealing with the noise present in our tubelets as described in Sec.~\ref{sec:mil_noise}. 
Note that for mean and LSE-pooling, we set $r = 1$. 
Finally, introducing our uncertainty-based loss function improves results even further, obtaining a Video mAP of 35.0 at a threshold of 0.5.
This is 80\% of the performance achieved by our fully-supervised baseline.

\paragraph{Person detections on UCF101-24}
Note that for our weakly-supervised experiments, the person tubelets for training are obtained from a Faster-RCNN~\cite{ren_neurips_2015} person detector that has only been trained on Microsoft COCO~\cite{lin_eccv_2014}.
There is a significant domain gap between COCO and UCF, and the annotation protocol of person boxes on UCF is also not consistent (for example, the bounding box for a person riding a horse often includes the horse in UCF) with that of COCO.
These discrepancies are reflected by the fact that our person detections used during training only have a recall of 46.9\% compared to the ground truth person boxes, when using an IoU threshold of 0.5 to signify a correct match.
Furthermore, the precision of our person tubelets on the training set is only 21.1\%.
A major contributing factor to this is that UCF action annotations are not exhaustive -- there may be people in the video who are not labelled at all as they are not performing an annotated action.
These people will, however, still be detected by a COCO-trained detector and considered as false positives during this evaluation.

The fact that we are able to train our model with these annotations demonstrates the ability of our multiple instance learning method to handle label noise in the training set.
The inconsistencies in the UCF101-24 dataset labelling are detailed further in the supplementary, and has also been noted previously by Ch\'eron \etal\cite{cheron_neurips_2018}.

Noise in the person detections are not a problem for the training of our fully-supervised baseline, as it is trained with ground-truth boxes in addition to predicted boxes.
As we have box-level supervision in this case, predicted detections which have an IoU of more than 0.5 with a ground-truth detection are assigned the label of the ground-truth box, or the negative label otherwise, during fully-supervised training.

As the goal of this paper is not to develop a better human detector or tracker for building the person tubelets, we use the Faster-RCNN detector released publicly by Ch\'eron~\etal\cite{cheron_neurips_2018} for all our evaluations on the UCF101-24 validation set. 
This detector was originally trained on COCO and then finetuned on the UCF101-24 training set using Detectron~\cite{Detectron2018}.

\paragraph{The effect of tubelet sampling}
For the tubelets of length $K =16$ that we use, there is a mean of 33.1 tubelets per video in the UCF101-24 dataset. In computing this, we only consider tubelets which have a spatio-temporal IoU of less than 0.5 with each other. More tubelets would be obtained if we counted one from each frame of the video.

As we can fit a maximum of 16 tubelets onto a 16GB Nvidia V100 GPU, it is clear that it is necessary to sample the tubelets in each bag.
Note that UCF videos often have a high number of tubelets, as there are often many people in the video who are not labelled as performing an action.
As described in the previous subsection, this is also a significant source of noise.

\input{tables/ucf_batch_size}

Table~\ref{tab:ucf_batch_size} shows the effect of changing the batch size (number of bags), and the number of tubelets sampled per bag, such that GPU memory usage is maximised.
We can see that the uncertainty loss helps in all cases and that
accuracy decreases with low batch sizes. We believe this is due to batch normalisation statistics being too correlated when more tubes are from the same video.

\paragraph{Comparison to state-of-the-art}
\input{tables/ucf101_24_sota_comparison.tex}

Table~\ref{tab:ucf_sota_comparison} compares our results to the state-of-the-art.
The bottom-half of the table shows that we outperform previous weakly-supervised methods by a large margin.
The top-half shows that our fully-supervised baseline is also competitive with the fully-supervised state-of-the-art, although that is not the main goal of this work.
The fully-supervised methods which outperform our method are based on action detectors which directly predict the person proposals with the network, and are thus able to handle the person annotation peculiarities of the UCF101-24 dataset more effectively.
We do not observe any issues with person detections for our experiments on AVA in the next section. %

\paragraph{Qualitative Results}
\input{figures/ucf_qualitative}

Figure~\ref{fig:ucf_qualitative} presents qualitative results of our method.
The first two rows show success cases of our method where the tubelet detection and linking have performed well.
The third row shows a failure case, since the basketball player represented by the green track is not actually performing the ``Basketball Dunk'' action.
According to the UCF101-24 annotations, only the player represented with the blue track is performing this action.
This video clip is thus an example of a video where there are many people not performing the action annotated for the video, and is especially challenging for our weakly-supervised method.
The fourth row shows a different failure case as an error by the online tubelet linking algorithm (we used the same method as~\cite{kalogeiton_iccv_2017}) has made the identities of the two cyclists change after they occluded each other.

\subsection{Experiments on AVA}

In this section, we report what to our knowledge are the first weakly-supervised action detection experiments on AVA~\cite{gu2018ava}.
The AVA dataset labels keyframes in a 15-minute video clip, where each keyframe is sampled every second (\ie at 1 Hz).
The evaluation protocol of the AVA dataset measures the ability of an action detection model to classify the actions occuring in a keyframe given the temporal context around it.

We control the difficulty of the weakly-supervised action recognition problem by combining the annotations from $N$ consecutive keyframes into a single, clip-level annotation.
This effectively means that we are obtaining clip-level annotations for sub-clips of $N$ seconds from the original AVA video.
The weakly-supervised problem gets more difficult as $N$ increases, as the sub-clips get longer and the number of observed labels within each sub-clip increases.
Note that when $N = 1$, only the spatial localisation ability of the model is being tested, as during training, it is unknown which of the subclip-level labels correspond to each person tubelet in the MIL bag.
When $N > 1$, the subclip-level labels can correspond to zero, one or many of the person tubelets at different keyframes in the clip, and it is thus a more difficult task.
As an AVA video clip consists of 900 seconds, $N = 900$ represents the most extreme case when spatio-temporal annotations are discarded for the entire 15-minute video.

\input{tables/ava_results}
\input{tables/ava_sota_results}
\input{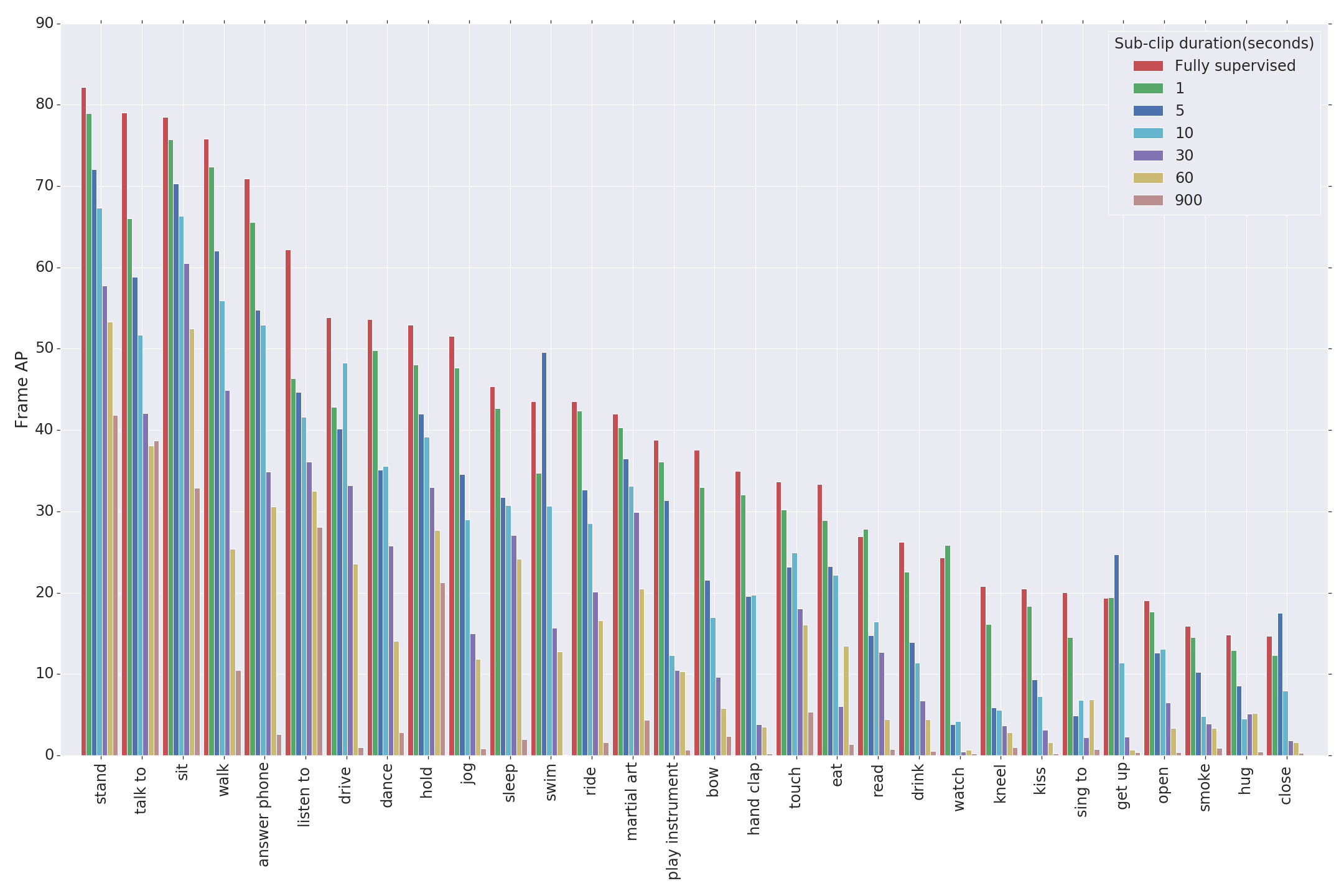}

Table~\ref{tab:ava_results} shows the results of our model in this setting.
As expected, the performance of our method improves the shorter the sub-clip.
For $N = 1$ and $N = 5$, our method obtains 90\% and 72\% of fully-supervised performance respectively, suggesting that bounding-box level annotations are not required for training action recognition models if the video clips are annotated over short temporal intervals.
Understandably, the results from $N = 900$, where we use the video-level annotations over the whole 15-minute clip are the worst as it is the most difficult setting.

Figure~\ref{fig:ava_per_class_results} further analyses the per-class results for the different levels of supervision presented in Tab.~\ref{tab:ava_results}.
As expected, stronger levels of supervision (shorter sub-clip durations) result in better per-class accuracy.
However, some action classes are affected more than others by weaker labels (longer sub-clips).
Examples of this include ``sing to'' and ``listen to'' which show a larger difference to the fully-supervised baseline than other classes.
Moreover, some classes such as ``watch (a person)'', ``get up'', ``close (e.g., a door, a box)'' and ``hand clap'' perform reasonably when trained with sub-clips ($N \leq 10$), but much more poorly when trained with longer sub-clips.

Finally, we compare our fully-supervised baseline to the state-of-the-art in Tab.~\ref{tab:ava_sota}.
Note that our weakly-supervised result from sub-clips of 10 seconds (Tab.~\ref{tab:ava_results}) outperforms the original fully-supervised baseline using introduced by the AVA dataset~\cite{gu2018ava} that uses both RGB and optical flow as inputs.
Our model, on the other hand, only uses RGB as its input modality.
Our SlowFast model performs similarly to the published results of the original authors~\cite{feichtenhofer_iccv_2019}.
Note that we have not used Non-local~\cite{wang_cvpr_2018}, test-time augmentation or ensembling which are all complementary methods to improve performance~\cite{feichtenhofer_iccv_2019}.
We can see that in contrast to the UCF dataset in the previous section, our person detector is accurate on AVA, and so a Fast-RCNN-style detector using person tubelets as proposals can achieve state-of-the-art results.

%% file: tables/ucf_ablation.tex
\begin{table}[t]
	\caption{Ablation study of different variants of our method on the UCF101-24 validation set. We report the Video mAP at IoU thresholds of 0.2 and 0.5 respectively.}
	\centering
	\begin{tabularx}{0.55\linewidth}{lYY}
		\toprule
		& \multicolumn{2}{c}{Video AP} \\
		& 0.2           & 0.5           \\ \midrule
		Weakly supervised baseline      &    54.3       &  29.7         \\
		MIL - LSE pooling               &    60.1       &  33.1         \\
		MIL - mean pooling           	&    60.3       &  33.0         \\
		MIL - max pooling               &    60.7       &  33.5         \\
		MIL - max pooling, uncertainty  &    61.7       &  35.0         \\ \midrule
		Fully supervised                &    69.3       &  43.6         \\ \bottomrule
	\end{tabularx}
	\label{tab:ucf_ablation}
\end{table}

%% file: tables/ucf_batch_size.tex
\begin{table}[t]
\caption{The effect of the number of bags in each training batch on accuracy (Video AP at 0.5). 
The uncertainty loss improves accuracy in all scenarios.
Although fewer, but larger, bags can reduce the noise due to sampling, they also cause batch normalisation statistics to be too correlated, reducing accuracy.}
\scalebox{0.66}{
\begin{tabularx}{1.5\linewidth}{YYYY}
\toprule
Number of bags in batch & Tubelets sampled per bag & Video AP without uncertainty & Video AP with uncertainty \\ \midrule
4                       & 4                        & 33.5            			  & 35.0            \\
3                       & 5                        & 33.6            			  & 34.1            \\
2                       & 8                        & 33.3            			  & 34.2            \\
1                       & 16                       & 25.8            			  & 26.2            \\ \bottomrule
\end{tabularx}
}
\label{tab:ucf_batch_size}
\end{table}

%% file: tables/ucf101_24_sota_comparison.tex
\begin{table}[t]
	\caption{Comparison to state-of-the-art methods on the UCF101-24 dataset in both fully- and weakly-supervised scenarios.}
	\centering
	\begin{tabularx}{0.7\linewidth}{lYY}
		\toprule
		& Video AP at 0.2           & Video AP at 0.5           \\ \midrule
	\multicolumn{3}{l}{\textit{Fully supervised}} \\
	Peng \etal\cite{peng_eccv_2016}	    & 42.3 & 35.9 \\
	Hou \etal\cite{hou_iccv_2017}			&  47.1 & -- \\
	Weinzaepfel \etal\cite{weinzaepfel_arxiv_2016}  & 58.9 & --                            \\
	Saha \etal\cite{saha_bmvc_2016}		 & 63.1 & 33.1 \\
	Singh \etal\cite{singh_iccv_2017}   	&  73.5             & 46.3              \\
	Zhao \etal\cite{zhao_cvpr_2019} & 78.5 & 50.3 \\
	Singh \etal\cite{singh_accv_2018}   	                             &  79.0             & 50.9              \\
	Kalogeiton \etal\cite{kalogeiton_iccv_2017}	 		             &   77.2        &  51.4          \\ 
	Ours 				   & 69.3 & 43.6     \\ %

	\midrule
	\multicolumn{3}{l}{\textit{Weakly supervised}} \\
	Escorcia \etal\cite{escorcia_arxiv_2018} 	&  45.5 & -- \\
	Ch\'eron \etal\cite{cheron_neurips_2018} 	  		                     &     43.9      & 17.7 \\
	Ours 	 			 & 61.7 & 35.0 \\  %
	\bottomrule
	\end{tabularx}
	\label{tab:ucf_sota_comparison}
\end{table}

%% file: figures/ucf_qualitative.tex
\begin{figure*}[t]

\centering
\def \imwidth {0.90\linewidth}
\newcommand \imagedir{figures/ucf_results/mil10_tubes/v_Fencing_g01_c05}
\newcommand \imageid {v_Fencing_g01_c05}

\begin{tabularx}{\linewidth}{YYYYY}
	\includegraphics[width=\imwidth]{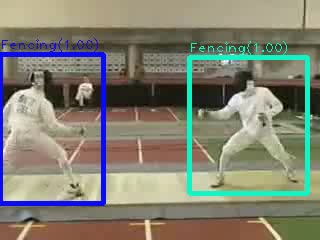} & 
	\includegraphics[width=\imwidth]{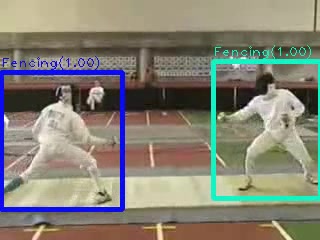} & 
	\includegraphics[width=\imwidth]{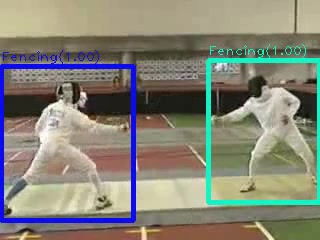} & 
	\includegraphics[width=\imwidth]{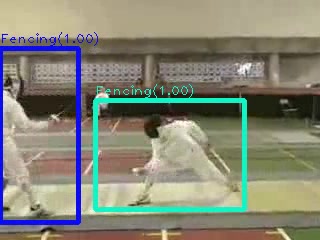} \\

	\includegraphics[width=\imwidth]{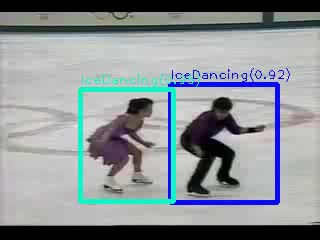} & 
	\includegraphics[width=\imwidth]{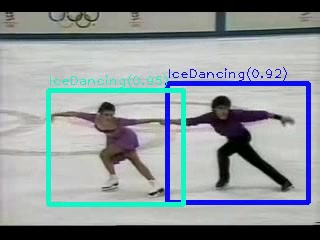} & 
	\includegraphics[width=\imwidth]{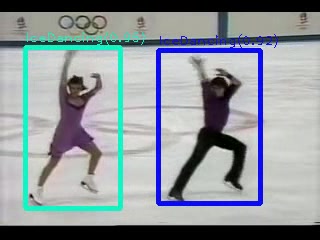} & 
	\includegraphics[width=\imwidth]{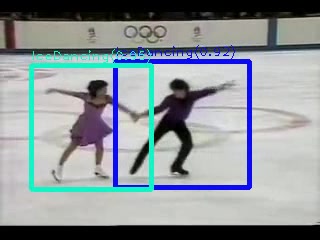} \\ 
	
	\includegraphics[width=\imwidth]{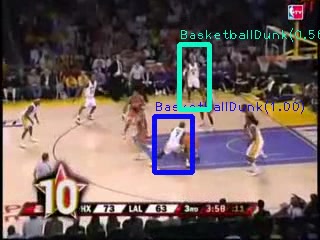} & 
	\includegraphics[width=\imwidth]{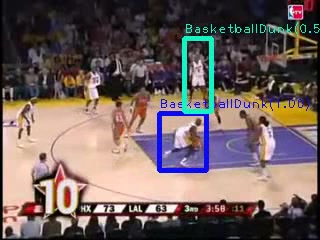} & 
	\includegraphics[width=\imwidth]{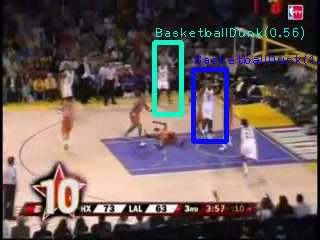} & 
	\includegraphics[width=\imwidth]{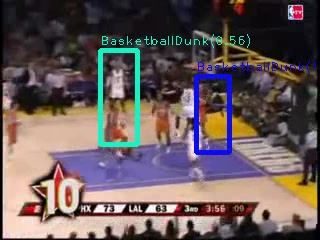} \\ 
	
	\includegraphics[width=\imwidth]{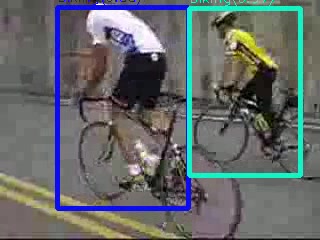} & 
	\includegraphics[width=\imwidth]{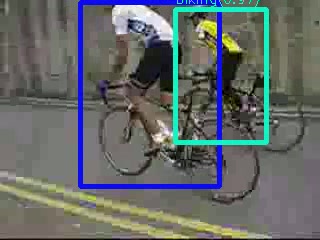} & 
	\includegraphics[width=\imwidth]{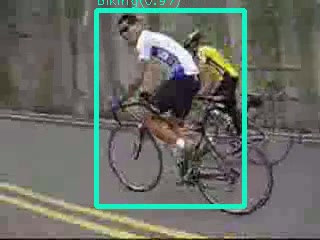} & 
	\includegraphics[width=\imwidth]{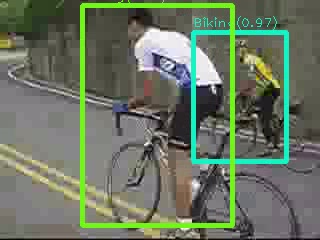} \\ 
	
\end{tabularx}
\caption{
Qualitative examples on UCF101-24.
Note that the bounding boxes are coloured according to the identity of the track. The action label, and tube score are labelled from the top-left of the bounding box.
Further discussion is included in the text.
}
\label{fig:ucf_qualitative}
\end{figure*}

%% file: tables/ava_results.tex
\begin{table}[tb]
	\centering
	\caption{
	Results of our method on the AVA dataset in terms of the Frame mAP at an IoU threshold of 0.5.
	We vary the length of the sub-clips from which we extract clip-level annotations to control the difficulty of the weakly supervised problems.
	FS denotes a fully-supervised baseline representing the upper bound on performance.
	A sub-clip of 900 seconds is an entire AVA video clip.
	}
	\begin{tabularx}{0.7\linewidth}{lYYYYYYY}
		\toprule
		& \multicolumn{7}{c}{Sub-clip duration (seconds)} \\ \midrule
				 				 & FS     & 1  		& 5  	 & 10  	   & 30  	 & 60  		& 900  \\
		Frame AP \hspace{0.05cm} &  24.9  & 22.4    & 18.0   & 15.8    & 11.4    & 9.1      & 4.2 \\ \bottomrule
	\end{tabularx}
	\label{tab:ava_results}
\end{table}

%% file: tables/ava_sota_results.tex
\begin{table}[t]
	\centering
	\caption{State-of-the-art fully-supervised methods on the AVA dataset.}
	\begin{tabular}{lc}
	\toprule
	Method & Frame AP \\
	\midrule
	AVA (with optical flow) \cite{gu2018ava} 						& 15.6 \\
	ARCN (with optical  flow) \cite{sun_eccv_2018}					& 17.4 \\
	Action Transformer \cite{girdhar_cvpr_2019}				& 25.0 \\
	SlowFast (ResNet 101) \cite{feichtenhofer_iccv_2019} 	& 26.8 \\
	SlowFast (ResNet 50, Ours)  	 						& 24.9 \\                            
	\bottomrule
	\end{tabular}
	\label{tab:ava_sota}
\end{table}

%% file: figures/ava_per_class_results.tex
\begin{figure}[th]
	\centering
	\includegraphics[width=0.98\linewidth]{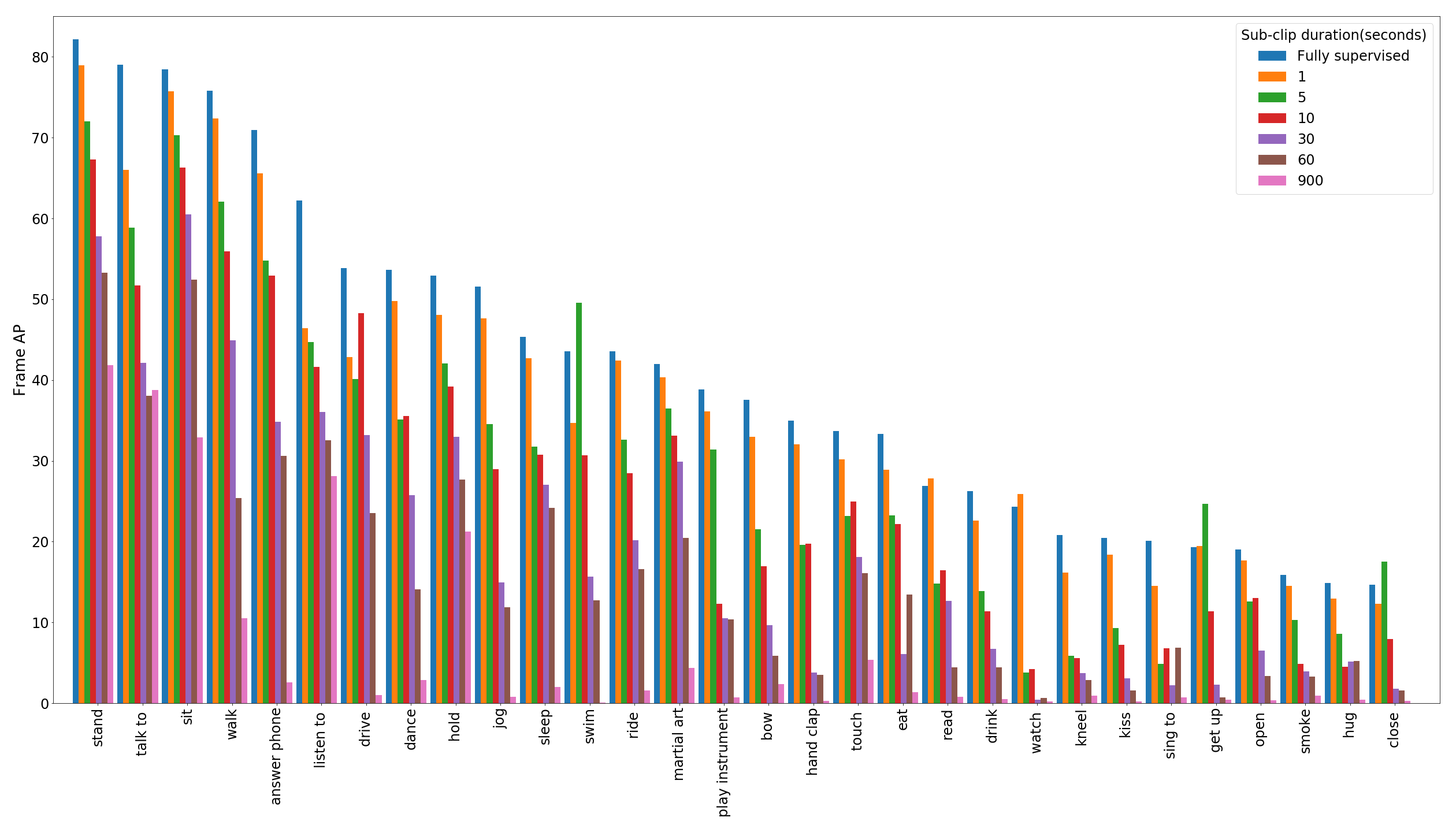}
	\vspace{-\baselineskip}
	\caption{Per-class results, in terms of the Frame AP, on the AVA dataset under different levels of supervision (the longer the sub-clip duration, the weaker the supervision). For clarity, the first 30 classes, ranked using the fully-supervised model, are shown.
	As expected, action classes benefit from stronger supervision, while some classes, such as ``watch'', ``get up'' and ``close'' are very difficult to learn from long sub-clips.
	}
	\label{fig:ava_per_class_results}
\end{figure}

%% file: text/conclusion.tex
\section{Conclusion and Future Work}
We have proposed a weakly supervised spatio-temporal action detection method based on Multiple Instance Learning (MIL).
Our approach incorporates uncertainty predictions made by the network such that it can better handle noise in our bags and violations of the standard MIL assumption by predicting a high uncertainty for noisy bags which cannot be classified correctly.
We achieve state-of-the-art results among weakly supervised methods on the UCF101-24 dataset, and also report the first weakly-supervised results on AVA, which is the only large-scale action recognition dataset.
Our analysis of the accuracy trade-offs as the time intervals for which sub-clips of the video are annotated will also aid future dataset annotation efforts.

Future work is to incorporate additional sources of noisy, weakly-labelled data, such as data which can be scraped off internet search engines.